\documentclass{article}

\usepackage{arxiv}

\usepackage[utf8]{inputenc} % allow utf-8 input
\usepackage[T1]{fontenc}    % use 8-bit T1 fonts
\usepackage{hyperref}       % hyperlinks
\usepackage{url}            % simple URL typesetting
\usepackage{booktabs}       % professional-quality tables
\usepackage{amsfonts}       % blackboard math symbols
\usepackage{nicefrac}       % compact symbols for 1/2, etc.
\usepackage{microtype}      % microtypography

\usepackage{microtype}
\usepackage{graphicx}
\usepackage{subfigure}

\usepackage{amsmath}
\usepackage{authblk}

\title{Explaining Motion Relevance for Activity Recognition in Video Deep Learning Models}
\author[1,2]{Liam Hiley}
\author[1,2]{Alun Preece}
\author[1,3]{Yulia Hicks}
\author[4]{Supriyo Chakraborty}
\author[6]{Prudhvi Gurram}
\author[5]{Richard Tomsett}

\affil[1]{Crime and Security Research Institute, Cardiff University}
\affil[2]{School of Computer Science, Cardiff University}
\affil[3]{School of engineering, Cardiff University}
\affil[4]{IBM TJ Watson Research Centre, IBM}
\affil[5]{IBM Hursley, IBM}
\affil[6]{US Army Research Laboratory, Adelphi}
\date{}
\begin{document}

\twocolumn[

% \keywords{Machine Learning, XAI, Explainable, Interpretable, Interpretability, Explainability, Activity Recognition, 3D CNN, Optical Flow}
\maketitle

\vskip 0.3in
]


\begin{abstract}
    A small subset of explainability techniques developed initially for image recognition models has recently been applied for interpretability of 3D Convolutional Neural Network models in activity recognition tasks. Much like the models themselves, the techniques require little or no modification to be compatible with 3D inputs. However, these explanation techniques regard spatial and temporal information jointly. Therefore, using such explanation techniques, a user cannot explicitly distinguish the role of motion in a 3D model's decision. In fact, it has been shown that these models do not appropriately factor motion information into their decision. We propose a \emph{selective relevance} method for adapting the 2D explanation techniques to provide motion-specific explanations, better aligning them with the human understanding of motion as conceptually separate from static spatial features. We demonstrate the utility of our method in conjunction with several widely-used 2D explanation methods, and show that it improves explanation selectivity for motion. Our results show that the selective relevance method can not only provide insight on the role played by motion in the model's decision --- in effect, revealing and quantifying the model's spatial bias --- but the method also simplifies the resulting explanations for human consumption.
\end{abstract}

\section{Introduction}\label{introduction}

\begin{figure*}[h]
    \centering
    \includegraphics[width=1.0\textwidth]{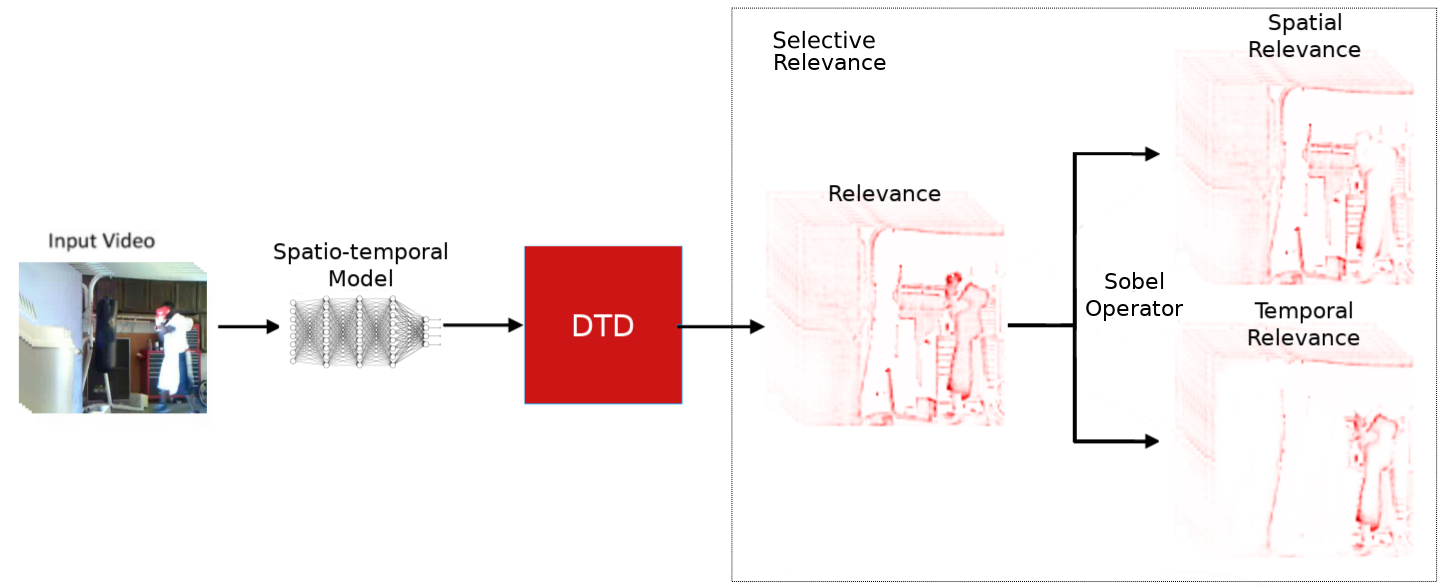}
    \caption{A step-by-step illustration of our approach for generating selective relevance maps for an explanation.}
    \label{fig:concept}
\end{figure*}

Among deep learning models for video activity recognition, 3D convolutional neural networks (CNNs)  are currently some of the most successful \cite{stroud:d3d,tran:spatiotemporal,carreira:quo,ji:3dconv,kensho-et-al:3dresnet,feichtenhofer:slowfast,diba:dynamonet}. These models rely on complex spatio-temporal feature extractors that take stacks of RGB frames as input, forming 3D cubes. By taking advantage of the convolutional kernels' ability to learn edges and other patterns, the model builds a representation of motion as edges in the third, temporal dimension. Treating motion as shapes and textures in a third dimension is a computationally efficient way to model the spatio-temporal features necessary for classification and recognition tasks; however, it is qualitatively different to the human understanding of speed, orientation and rhythm of motion when recognising movements such as swinging a racquet in tennis. This is further brought to light when the decisions made by these models are explained using widely-used 2D saliency map methods, from the image recognition domain, and extended to 3D domain~\cite{baehrens:sensitivity,simonyan:conv_visualising,zhou:cam,erhan:visualizing,selvaraju:gradcam,hooker:roar,montavon:deeptaylor}. While the saliency maps can effectively quantify the contribution of a spatio-temporal region to activity recognition, the model's utilization of motion information in making such decisions is not properly quantified and distinguished from that of spatial information. 

The extended 2D saliency maps hinder the human user's ability to interpret the explanation for a task that is inherently temporal such as activity recognition. They also obscure the role that motion plays in the model's decision, which may be crucial in evaluating and improving its performance. Moreover, the notion that 3D CNNs inherently do not always build adequate representations of motion has previously been explored and addressed in \cite{stroud:d3d,diba:dynamonet,feichtenhofer:slowfast}. In such cases, it is critical for saliency maps to accurately explain if the 3D CNN models are using motion information or just contextual/spatial information to recognize a particular activity.

In this work, we introduce the \textit{selective relevance} method for extracting the temporal component of a spatio-temporal explanation, generated via the above methods, for a basic 3D CNN activity recognition model (illustrated in Figure~\ref{fig:concept}). In doing so, we quantify the contribution of motion to the model's decision, and therefore reveal the degree to which the model is biased towards contextual, i.e. spatial, information. As a byproduct, our method produces simplified explanations that benefit from the removal of a large amount of relevance attributed to the background spatial features. We evaluate our selective relevance method by using it to extend a subset of widely-used baseline explanation techniques and find that: (1) our method consistently improves the representation of motion in explanations over the baseline methods by removing areas of the explanations associated with little to no motion; (2) overall, the baselines agree that motion plays a small role in our model's decisions; and (3) the selective relevance method can be added to these baseline explanation techniques for only a small additional computational cost.

\section{Related Work}\label{related}
\subsection{CNNs for Activity Recognition}
Originally, video problems such as activity recognition were tackled using image recognition techniques by processing each RGB frame individually as a stand-alone image \cite{karpathy:large}. As such, models could only infer decisions based on spatial, or `contextual' cues (e.g., a tennis racquet, surf board, etc). In order to incorporate the intuitively-necessary temporal motion information into the model decision, many such 2D CNNs were folded into recurrent networks \cite{donahue:long_rnn}. Along the same lines, inflating 2D convolutional kernels of the network to three dimensions was first explored in \cite{ji:3dconv,kensho-et-al:3dresnet}. This leads to a finer-grained representation of motion as third dimensional shapes and textures, that does not rely on sequences of spatial cues so much as combined spatio-temporal features. Seeking to benefit from motion information provided by optical flow, in \cite{carreira:quo}, the authors used two-stream approach \cite{simonyan_zisserman:two-stream,wang:sr-cnn,wang:temporal} and fused two 3D CNNs, one stream using original video frames and the other using optical flow between the frames, achieving state-of-the-art activity recognition performance. However, this highlighted a key issue with the manner in which 3D CNNs currently learn motion representations: intuitively, with all the information presented, a 3D CNN should capably learn relationships between pixels in the third dimension as temporal features. Yet the addition of optical flow, which quantifies this relationship, has consistently proven to improve the model's predictive power. This lack of motion understanding in 3D CNNs was addressed in works such as D3D and DynamoNET \cite{stroud:d3d,diba:dynamonet}, which implement additional steps specifically geared towards distilling motion representation in the model during the training phase, and in SlowFast Networks, which experiment with various spatial temporal resolutions \cite{feichtenhofer:slowfast}. The SlowFast model operates on two feature streams, both from 3D RGB inputs. However, while one stream takes the full resolution frames, its temporal resolution is relatively low. The `fast' stream makes up for this by taking input over a much longer set of frames, but at a significantly downsampled spatial resolution. Through this multi-resolution approach the authors were able to obtain the current state-of-the-art in action recognition performance. In this paper, we propose a method to generate motion-specific saliency maps that would explain if a 3D CNN-based activity recognition model is, in fact, using motion information to make decisions.

\subsection{Explainable Deep Learning}

A large body of work has been carried out in explaining deep learning models, particularly in the image recognition field. Many of these methods estimate the relevance/contribution of the different input features (i.e., pixels) to a model's output, representing this relevance as a heatmap over regions of the input \cite{baehrens:sensitivity,zhou:cam,selvaraju:gradcam,bach:lrp,montavon:deeptaylor}. The distinction between these techniques is found in how this contribution is defined and distributed mathematically. Many techniques use backpropagation along the model graph from the output to the input, an approach first formalised in \cite{baehrens:sensitivity}. Different methods use different rules to constrain this signal.

Explanations following the Layer-wise Relevance Propagation (LRP) approach \cite{bach:lrp}, most notably \cite{montavon:deeptaylor,zhang:eb}, show a grey-scale pixelmap, where pixel intensity is proportional to that pixel's relevance to the decision. In \cite{zhang:eb}, the authors additionally demonstrate explaining the dual (all other classes to the target class) to the target class in a process, they call Contrastive Marginal Winning Probability or cMWP. This second explanation is useful as destructive noise for the first, suppressing relevance that is common to all classes. In \cite{montavon:deeptaylor}, the Deep Taylor Decomposition (DTD) method additionally defines bounds within which to search for root points (a property of LRP) based on the input. As such, DTD is able to produce crisp, more focused explanations that effectively highlight salient pixels.

Class Activation Mapping, or CAM \cite{zhou:cam}, upsamples the final feature map of Global Average Pooling (GAP) Networks to the input size to produce a heatmap of model attention. GAP networks do not flatten the output or use fully connected layers and as such retain the relative position of filter activations. GradCAM, \cite{selvaraju:gradcam}, extends this to regular CNNs that use fully connected layers, by finding the gradient of the output classification neuron with respect to each output position in the final feature map and thus quantifying the contribution of each subregion to the classification. The resulting map will then show the strength of contribution of each subregion to the classification similar to the original CAM. The most evident drawback of this method is that the coarseness of the heatmap is directly proportional to the degree of downsampling that occurs in the convolutional layers. To emulate the more fine-grained approach found in LRP and other backpropagation methods, GradCAM can be paired with Guided Backpropagation \cite{springenberg:gb}, to produce Guided GradCAM, a pixelwise quantification of contribution, influenced by the regional attention of GradCAM.

\subsection{Explainable Video Deep Learning}

While developed for 2D CNN models, these explanation methods are applicable to other input domains, as long as the differentiable nature of the network is upheld. Such an extension to video activity recognition models is explored in works such as \cite{srinivasan:videolrp,stergiou:tubes,hiley:discdtd}. In \cite{stergiou:tubes}, the authors apply GradCAM \cite{selvaraju:gradcam} to 3D CNNs. This results in the characteristic heatmap of the GradCAM method, over a number of frames, which when visualised appears as a tube of saliency moving along the temporal dimension. The work of \cite{srinivasan:videolrp} demonstrates the application of DTD to a compressed domain action recognition problem. Here, DTD is applied to a composite model consisting of an SVM applied to Fisher Vectors created from input frames. The explanation is displayed as key frames over time, measuring the distribution of relevance over time. Building upon these works \cite{srinivasan:videolrp,stergiou:tubes} and \cite{zhang:eb}, \cite{hiley:discdtd} demonstrates a method for extracting spatial and temporal components from 3D relevance. 

To our knowledge, \cite{hiley:discdtd} is the first study to explore the concept of temporal relevance. The authors approximate the two hidden components through padding: by explaining single frames inflated to input size, the authors suppress all temporal information from the explanation for that frame. As such the model could only have inferred its decision from spatial information. Similar to cMWP, this can be used to suppress spatially relevant regions in the original explanation via subtraction, which would then highlight the remaining temporally relevant regions. However, their solution exhibits some key weaknesses. First, it is only able to highlight regions of possibly high temporal relevance. However, the approach can not accurately decompose the relevance into its spatial and temporal components, as evident in the existence of negative relevance in the temporal explanations illustrated in \cite{hiley:discdtd}. Second, by inflating each frame to match the input shape, the authors create new inputs that the model has not seen before. This results in explanations that are not truly faithful to the model and therefore cannot be trusted. Finally, the computational overhead for generating the explanation, which involves multiple forward \emph{and} backward passes, grows linearly with the number of frames in the input. To overcome the above weaknesses, we propose a new method called selective relevance in the next section.

\section{Selective Relevance}\label{SR}

\begin{figure}[!]
    \centering
    \includegraphics[width=0.35\textwidth]{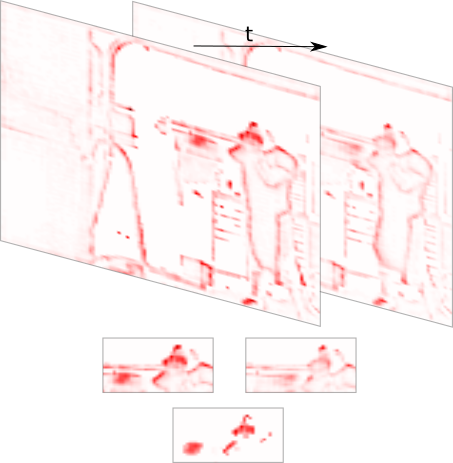}
    \caption{An example of the use of a Sobel filter in the $t$ axis, as a mask on a pair of frames from an explanation of the UCF-101 activity BoxingPunchingBag. The discrete derivative for the selected region is taken between $t$ and $t + 1$. Taking high values in the product as a mask over the original first frame results in the explanation as shown in the bottom image.}
    \label{fig:sobel}
\end{figure}

A selective relevance map is generated by decomposing an explanation for 3D spatio-temporal input into spatial and temporal components via derivative-based filtering, to discard regions with near-constant relevance over time. We propose to take the discrete derivative of the relevance in each of the three dimensions. By doing so we aim to select regions for which the relevance changes sharply over time. This represents information that the model considers relevant due to change that has occurred in that region in the neighbouring frames, causing the model to shift focus.

\emph{Capturing motion relevant to model decision:} Measuring the temporal derivative at a pixel level for the input video would result in a high value for any sharp movement, as the pixels in the region of the movement would likely significantly change in colour and intensity. To avoid this, we take the derivative of the explanation rather than that of the input video. The reasoning behind this is that, regardless of the severity of the motion, if the region is and remains irrelevant to the model decision, the pixel relevance value will remain near constant in the explanation. 

To obtain this discrete derivative for an image-like tensor, we employ Sobel edge-detection. The Sobel operator is a specific configuration of a convolutional kernel that when passed over an image, produces a grayscale intensity map of the edges in that image (as shown in Figure~\ref{fig:sobel}). The operator works in either the $x$ or $y$ dimension, after which the gradient magnitude can be taken to combine these. It can also be extended to the third dimension to detect `edges in time' in the explanation and generate a temporal edge map ($t$). We therefore approximate the pixel-wise temporal derivative $G_t$ of the relevance $R$ via convolution (here denoted by $*$) using the Sobel kernel, $h'_t$.
\begin{equation*}
    G_t(R) = h'_t * R
\end{equation*}

By applying a threshold value of $n$ standard deviations ($\sigma$), which is defined by the user requesting the explanation, we can produce a mask from the temporal edge map ($t$) by setting the voxels in the temporal edge map that are greater than the threshold, $\sigma$, to one, and zero otherwise. We apply this mask to the original baseline relevance provides a filtered temporal relevance map, which we call a selective relevance map. In other words, the generated mask is then used to extract temporally relevant regions from the explanation for a 3D input cube of size T~$\times$ H~$\times$ W, where T is the number of frames, and (H, W) is the size of each frame.
\begin{align*}
    &R_t = \{r_{ijk} | G_t(r_{ijk}) > \sigma\}\\
    &\forall i \in 1, \ldots, T, j \in 1, \ldots, H, k \in 1, \ldots, W
\end{align*}

\section{Experimental Setup}
\label{experiments}

We evaluate our selective relevance approach in terms of three criteria: \textit{motion selection}, i.e., precision with optical flow as ground truth; \textit{selectivity}, i.e. proportion of relevance in selective explanations vs their spatio-temporal baselines; and \textit{overhead} of running our selective relevance method on top of the baseline explanation process. We perform this analysis on a randomly-selected subset of 1,010 videos from the UCF-101 dataset.
For all selective explanations generated, we took a threshold of four standard deviations when filtering motion.

\subsection{Model}
We use a PyTorch implementation of the C3D 3D CNN \cite{tran:spatiotemporal} trained for activity recognition. We fine-tuned the model using initial weights trained on the Sports-1M \cite{karpathy:sports1m} dataset from \cite{tran:spatiotemporal}. These were ported to PyTorch from the original Keras implementation. For ease of use of the DTD method, and to avoid bringing in any discussion on additional LRP rules, we choose not to combine the CNN with other Machine Learning (ML) methods to improve the accuracy, as done with a bag-of-words approach for motion feature extraction in \cite{ji:3dconv}, and with a linear SVM for classification in \cite{tran:spatiotemporal}.

\subsection{Training}
We trained the model for 60K iterations on a batch size of 32 samples, using stochastic gradient descent with a starting learning rate of 1~$\times$ 10\textsuperscript{-3} decreasing by a factor of 0.1 after 10 epochs of stagnation. Samples were scaled and then center-cropped to 112~$\times$112, and to sixteen frames. In the event that the final cropping of the video was less than sixteen frames, we use loop padding. Additionally we zero-center the frames using the channel-wise means for UCF-101. The training was run on two 1080 Ti GPUs over 24 hours.
With these settings the model was able to achieve 74\% test accuracy on the dataset. We note that this is below state-of-the-art by a significant margin. However, in the interest of model interpretability rather than obtaining/improving state-of-the-art, we find this model sufficient for the application.

\subsection{Evaluation Criteria}
\subsubsection{Motion Selection}
It is not possible to evaluate the ``correctness" of the generated selective relevance maps directly, as there is no ground truth against which to compare our estimated maps. However, we can quantify the amount of relevance in our method's explanation that is attributed to motion, using the optical flow as a ground truth for motion. Optical Flow is a well-established technique for extracting approximate pixel motion from a pair of consecutive frames. In theory, any area of non-zero relevance in an explanation from our method should lie in an area of non-zero optical flow, in order for us to refer to it as relevant motion. Therefore, any temporal relevance that has a low corresponding optical flow value can be considered incorrectly assigned. This gives us a measure of the amount of motion relevance that is actually motion, allowing us to quantify the performance of our method vs. the baselines, without the need for a ground truth for saliency.

In order to attribute a pixel as `relevant' in an explanation, we center all explanations around zero, as before the various visualisation steps. As a result, all areas of negligible relevance are marked as zero and can be easily filtered out.

For a saliency map produced via our method, $R$, and a dense optical flow field $O$, we compute the voxelwise `precision', $P$ of the two as the percentage of voxels in which both relevance and optical flow are positive:
\begin{align*}
    &P = \frac{\sum\limits_{i}\sum\limits_{j}\sum\limits_{k}(I_R \odot I_O)}{\sum\limits_{i}\sum\limits_{j}\sum\limits_{k}(I_R)}\\
    &\text{Where } I_R = R > 0,
\end{align*}
and $\odot$ is the hadamard product.

Note that we cannot compute a `recall' measure because optical flow would be non-zero if there is camera motion, even when the motion related to a particular activity might be non-existent. Hence, it is not possible to calculate false negatives and thus, not possible to measure `recall'.

\subsubsection{Selectivity}
We propose that selective relevance can function as a filter for spatio-temporal explanations, removing much of the contextual/spatial relevance that is assigned to features in the background of the scene. We, therefore, measure the ratio of positions at which there is positive relevance in the selective explanation, with respect to the corresponding baseline explanation. We also measure the ratio of the sum of relevance, in both.

In this context, the ideal result would be that the majority proportion of relevance be present in a minority proportion of area, thus simplifying the explanation for human consumption without losing the representation of the model's decision.

\subsubsection{Overhead}
Real-time scenarios are common in video deep learning. We therefore evaluate our method's viability for a motion-specific explanation as an additional benefit, in the case of real-time evaluation of activity recognition decisions. We can not improve on the computation time of the baseline methods, so we time the overhead of selective relevance on top of the baselines in order to assess its viability in this context.

The chosen model takes inputs of consecutive 16 frame stacks, which, for a framerate of 30 fps is just over half a second. We use this as a loose upper bound for explanation time in order to minimise lag.

\section{Results}\label{results}
\begin{figure*}[t]
    \centering
    \includegraphics[width=1.0\textwidth]{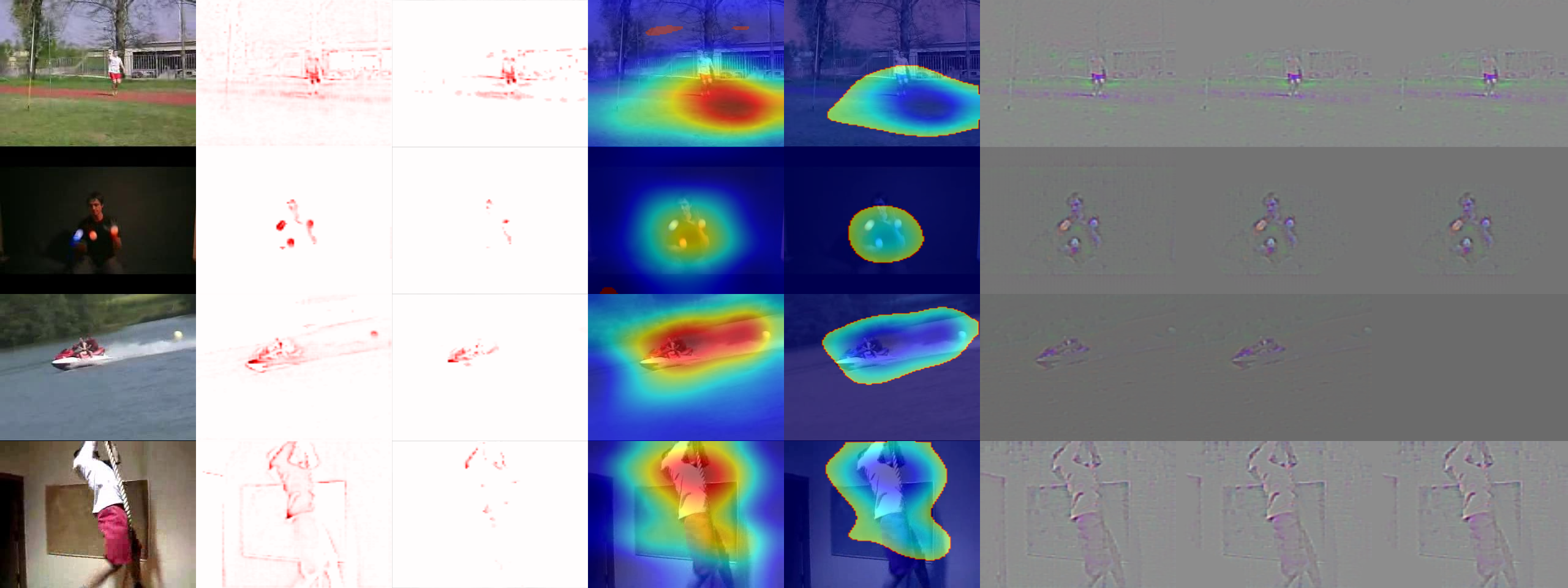}
    \caption{From top to bottom: samples from UCF-101 classes JavelinThrow, JugglingBalls, Skijet and RopeClimbing respectively. From Left to Right: Original frame for reference, 3D DTD explanation, Selective DTD, 3D GradCAM explanation, Selective GradCAM, 3D Guided Backprop explanation, 3D Guided GradCAM explanation, Selective Guided GradCAM.}
    \label{fig:sr}
\end{figure*}
As previously stated, we generated explanations for 1,010 videos, composed of 12,219 clips. Figure~\ref{fig:sr} shows illustrative examples from the top 4 samples as per our Motion Selection study. For comparative analysis, we present the results in eight columns, demonstrating the input frame, each of the three baseline explanations and their selective counterparts. We also include Guided Backpropagation which makes up the other half of Guided GradCAM.

\subsection{Motion Selection}
\begin{table}[ht]
    \centering
    \resizebox{0.45\textwidth}{!}{
    \begin{tabular}{ccc}
        \hline
        Method & Average (\%) & Standard Deviation (\%) \\
        \hline
        DTD & 27.79 & \textbf{16.72} \\
        % \hline
        Selective DTD & 41.09 & 18.89 \\
        % \hline
        GradCAM & 28.17 & 16.78 \\
        % \hline
        Selective GradCAM & 36.12 & 17.11 \\
        % \hline
        Guided Backpropagation & \textbf{47.48} & 17.44\\
        % \hline
        Guided GradCAM & 42.89 & 17.03 \\
        % \hline
        Selective Guided GradCAM & 45.76 & 17.12\\
        \hline
    \end{tabular}}
    \caption{Precision is the percentage overlap of pixels with positive relevance in the explanation and pixels with positive flow in the optical flow field.}
    \label{tab:flow}
\end{table}
In this section we demonstrate the advantages of the selective relevance method for the various corresponding baselines, over the original, full explanations. As might be expected, the diverse explanations generated by the three baselines result in different behaviours when our selective approach is applied.

For instance, DTD exhibits low precision for explanations, often finding some degree of relevance in all objects in the scene. In the best case, where the scene is blank, as can be found in the juggling example, DTD still displays the juggler's head and all three balls as relevant. In more common cases, this results in the `decisive' features being obscured by large strokes of relevance, as can be seen clearly in the rope climbing and javelin throw examples. By applying selective relevance, the resulting explanations are shown to be much more focused and simpler. We can see that while much of the background is considered relevant, almost all of it is actually static aside from the shoulders and upper torso, and the runner's legs and the track respectively. Most clear is the exclusion of the surf and the buoys in the jet ski example, leaving only the jet itself, despite all of the above being in motion.

In the case of GradCAM our method does little to improve the highly contextual explanations. Likely due to the upsampling step, there is actually very little change of saliency over time. Still, we can see that the centre of focus remains quite stably relevant for most explanations, whereas the edges fluctuate much more strongly, with red areas in the Selective GradCAM explanations representing higher intensity change.

When combined with Guided Backprop, GradCAM produces focused, pixel-level salience maps. As can be seen in all examples, background relevance falling outside of the GradCAM heatmap is suppressed. Selective Guided GradCAM displays the same properties, but as a result also suffers from the same issues as Selective GradCAM.

Quantitatively (Table~\ref{tab:flow}), Guided Backpropagation scores highest in precision on average, closely followed by Guided and Selective Guided GradCAM. Guided Backpropagation generally provides focused, fine-grained explanations, explaining the high score. One would therefore expect that such a fine-grained, feature-level saliency map, weighted by the regional focus of GradCAM, and filtered temporally would produce the most faithful temporal explanation. Overall DTD alone is the worst performing as expected due to its heavy sensitivity to contextual information. Our method applied to DTD is very close on average, demonstrating that our method is capable of focusing the noisy DTD explanations down to the level of Guided methods' explanations. GradCAM performs similarly poorly, even with the addition of the selective step, perhaps suggesting that it is heavily contextually oriented. As stated before, the final feature map before up-sampling is low-resolution, in our case being a 2$\times$ 7$\times$ 7 map, here there are two temporal regions and therefore temporal focus is very coarse-grained at eight frames per region.

\subsection{Selectivity}
\begin{table}[h]
    \centering
    \resizebox{0.45\textwidth}{!}{
        \begin{tabular}{ccccc}
        \hline
        % \multirow{2}{*}{}
        &
          \multicolumn{2}{c}{GradCAM} &
          \multicolumn{2}{c}{Guided Backpropagation} \\
        \cline{2-3} \cline{4-5}
        & Avg. & Std. & Avg. & Std. \\
        \hline
        DTD & 36.61\% & 13.84\% & 2.79\% & 1.78\%\\
        GradCAM & - & - & 0.95\% & 0.48\%\\
        \hline
        \end{tabular}}
    \caption{A bitwise comparison of the allocation of relevance in the various baselines we use. The percentage agreement shown here is the percentage overlap between areas of relevance in explanations from two baselines.}
    \label{tab:comp}
\end{table}
\begin{table}[t!]
    \centering
    \resizebox{0.45\textwidth}{!}{
        \begin{tabular}{ccccccc}
        \hline
        % \multirow{2}{*}{} &
            &
          \multicolumn{2}{c}{DTD} &
          \multicolumn{2}{c}{GradCAM} &
          \multicolumn{2}{c}{G. GradCAM} \\
        \cline{2-7}
        & Avg. & Std. & Avg. & Std. & Avg. & Std. \\
        \hline
        Our method & 7.7 & 4.1 & 4.7 & 0.7 & 41.9 & 7.8 \\
        & (10.5) & (7.1) & (2.6) & (0.3) & (36.9) & (7.6)\\ 
        \hline
        \end{tabular}}
    \caption{A measure of selectivity through the same process as tab. \ref{tab:comp}. Shown are the percentage overlaps between a baseline and it's Selective counterpart. The selective explanation through our method selects a fraction of the original explanation, which we quantify here.}
    \label{tab:selection}
\end{table}
In order to assess our method for selecting motion relevance, we consider it important to assert that these baselines agree to some extent on the relevance in general. For instance, while Guided Backpropagation is by far the most precise in assigning relevance to motion, it disagrees drastically with both DTD and GradCAM in Table~\ref{tab:comp}. This suggests it is also possible that the high performance is instead due to a poor representation of the model's attention, with the majority vote being that the model is much more biased towards context.

Quantitatively all baselines suggest a model bias for contextual information, as exhibited in the large proportion of relevance lost in the selective process (Table~\ref{tab:selection}). This is additionally useful in suggesting the earlier dubious nature of Guided Backpropagation: While the proportion of relevant pixels within motion is considerably higher for Guided GradCAM than GradCAM (as well as DTD), the proportion of relevance allocated is actually lower. I.e. while 42\% of relevant pixels are due to motion, this only represents 37\% of the relevance, suggesting that most of the models' attention is more heavily focused towards contextual information. This is more in agreement with DTD and GradCAM, which suggest that around 90\% of the information the model considers relevant is relatively stable over time, i.e. likely contextual. Overall it seems that while DTD is temporally dynamic, with selective explanations consistently exhibiting a larger fraction of the original relevance than that of the relevant pixels, GradCAM and Guided Backprop are very stable, and as such little relevance was selected as `temporal'. 

In order to produce better precision selective explanations, we suggest that the threshold ($\sigma$) in Section~\ref{SR} should be treated as a hyperparameter to be optimised for each dataset/task (in this case being lowered), rather than a constant. Experimentation is strongly encouraged to find the best fit.

\subsection{Overhead}
\begin{table}[h!]
    \centering
    \resizebox{0.45\textwidth}{!}{
        \begin{tabular}{ccccccc}
        \hline
        % \multirow{2}{*}{} 
        &
          \multicolumn{2}{c}{DTD} &
          \multicolumn{2}{c}{GradCAM} &
          \multicolumn{2}{c}{Guided GradCAM} \\
        \cline{2-7}
        & Avg. (ms)& Std. (ms)& Avg. (ms)& Std. (ms)& Avg. (ms)& Std. (ms)\\
        \hline
        Original & 1.16 & 0.455 & 136 & 3.19 & 246 & 3.51\\
        Selective & 1.71 & 0.582 & 137 & 3.19 & 247 & 3.52\\
        \hline
        \end{tabular}}
    \caption{For each baseline: DTD, GradCAM and G-GradCAM, we calculate the average time taken to generate an explanation for any given input video, and the variation between times (i.e. the standard deviation). We also do this for the Selective versions of these methods.}
    \label{tab:speed}
\end{table}
In this subsection, we explore the overhead for running selective relevance on top of the various baselines to show that the process is relatively cheap in exchange for obtaining a much simpler, motion-specific explanation. Our results are recorded in Table~\ref{tab:speed}.

Overall, baseline explanations are relatively quick to generate with access to the resources as described in Section~\ref{experiments}, with DTD on average executing in a millisecond, and GradCAM and Guided GradCAM both running in under half a second.
The Guided GradCAM explanations, being a combination of two separate explanation algorithms, takes the longest at $2.46\times10^{-1}$ s.

With the addition of the selective process, the overhead time to generate the explanation is negligible, with DTD suffering the greatest increase at 50\% (its computation time being at the level of the overhead). The increase in time appears to be a constant overhead of between 0.5 and 1 milliseconds. We therefore state the viability of our method in a real-time setting for negligible increase in lag. The availability of a near real-time, motion oriented explanation provides the user with additional insight to the model's perception of the current stream of events in relation to each other. Moreover, selective relevance's consistently simpler explanations would, most notably in the case of DTD, greatly assist an user in picking out key, relevant entities in the scene.

In order to qualify our results, we carried out some initial experiments on real-time selective explanations. We were able to provide near real-time explanations for an incoming video stream, e.g., from a live Webcam using our proposed approach. This proved to be very useful during development, not only for quick debugging of the method in a deployment scenario, but also for confirmation that our method is useful to developers as well as end users. In fact, we quickly picked up insights into the model's strengths and weaknesses with the addition of selective relevance. Understanding the inner workings of a model is one of the key motivations for the development of explanation methods, and a live heatmap showing the contribution of the motion to a model's decision is an effective way to qualify the approach.

\subsection{Discussion of Results}\label{conclusion}
For all explanations, background information such as trees, water and a chipboard are marked as relevant. Therefore, we view the removal of these objects from the explanation as a success for our method in improving the interpretability of the original explanation. Moreover, the remaining selective relevance in the examples was attributed to objects intuitively relevant to each activity for motion, such as  athletes limbs, or the vehicle. This is fortified by the increase in precision with regards to optical flow, for each of the three chosen baselines. The highest precision is produced by Guided Backpropagation, which creates highly focused, fine-grained explanations, but agrees very little with DTD and GradCAM in terms of relevance distribution.

Overall, DTD is shown to benefit the most from selective relevance, with the largest increase in precision, causing it to become comparable to GradCAM and Guided GradCAM, and qualitatively the greatest reduction of contextual clutter. While explanations from GradCAM and Guided GradCAM are also reduced significantly, this often results in far too much relevance being stripped away, and in many cases, most frames contain no selective relevance whatsoever.

Again, we believe that the choice of threshold for this method is data and use-case dependent, and most likely subjective. Our preference for a more selective explanation is influenced by the context and might not improve interpretability in other cases. In this case, a threshold of two standard deviations proves to be unhelpful with GradCAM and Guided GradCAM.

We show that there is little agreement between baselines in their assignment of relevance, though this is influenced by the coarse-grained explanations produced by GradCAM. Perhaps most interesting is the large disagreement between DTD and Guided Backpropagation, both being fine-grained, but DTD on average being much less focused. This provides an argument against the selection of Guided Backpropagation as the optimal method for explaining relevant motion.

In terms of selectivity, the results suggest that DTD and GradCAM are heavily biased towards contextual relevance. Their agreement in this respect strengthens the argument that the model itself is biased towards contextual information, and that both baselines are in fact more faithful than Guided Backprop and by extension Guided GradCAM. This strengthens the observation that 3D CNNs fail to effectively learn motion for activity recognition without additional steps. Both Guided baselines exhibit much higher agreement with motion, and as such selectivity is much less drastic. If contrarily the model is not biased towards context, this would suggest that either Guided Backpropagation, Guided GradCAM or Selective Guided GradCAM are the best at providing motion explanations.

\section{Conclusion and Future Work}\label{discussion}
We have introduced selective relevance, a post-processing step for pixel-relevance based explanation methods applied to 3D CNNs for video classification. selective relevance separates the temporal and spatial components of relevance maps, improving the comprehensibility of the explanations by better aligning them with humans' intuitive understanding of time and space as separate concepts. Importantly for many video tasks, the method is simple and fast: we have used selective relevance ourselves to provide real-time explanations during model development and debugging.

selective relevance consistently improves the proportion of motion in explanations on the baseline, providing a distilled, visually simpler representation of the model's decision process for tasks that are inherently temporal in nature such as activity recognition. It also provides insights into the model's bias for contextual spatial information when classifying a video: selective relevance revealed that our 3D CNN was heavily biased towards contextual spatial information, an observation backed up by all of our baselines.

Through this efficient post-processing step, we have been able to identify an issue within a learned model's understanding, without the need for analysis of training or performance. While it has been previously established that these models need additional attention in learning motion within the context of activity, we have not only confirmed this using post-hoc explanation methods, but through doing so, effectively quantified the extent of the bias.

The next logical step in this work is to perform a similar study to a more recent, state-of-the-art model such as D3D, with the hope of finding significantly improved performance as per these measures. In the D3D architecture, the filters of the model are specifically trained to emulate optical flow based features, and as such we would expect relevance for these filters to be much more in agreement with the actual optical flow for the video.

\section*{Acknowledgements}
This research was sponsored by the U.S. Army Research Laboratory and the U.K. Ministry of Defence under Agreement Number W911NF-16-3-0001. The views and conclusions contained in this document are those of the authors and should not be interpreted as representing the official policies, either expressed or implied, of the U.S. Army Research Laboratory, the U.S. Government, the U.K. Ministry of Defence or the U.K. Government. The U.S. and U.K. Governments are authorized to reproduce and distribute reprints for Government purposes notwithstanding any copyright notation hereon.

% \bibliography{main}
% \bibliographystyle{ieeetr}

\end{document}